\documentclass[british]{article}
\usepackage[T1]{fontenc}
\usepackage[utf8]{inputenc}
\usepackage{babel}
\usepackage{varioref}
\usepackage{textcomp}
\usepackage{tipa}
\usepackage{graphicx}

\makeatletter

\providecommand{\tabularnewline}{\\}

\usepackage{qtree}
\date{}

\makeatother

\begin{document}

\title{A human-editable Sign Language representation for software editing---and
a writing system?}

\author{Michael Filhol\\
\texttt{\footnotesize{}michael.filhol@limsi.fr}\\
{\footnotesize{}LIMSI, CNRS, Université Paris Saclay}\\
{\footnotesize{}Orsay, France}}

\maketitle

\begin{abstract}
To equip SL with software properly, we need an input system to represent
and manipulate signed contents in the same way that every day software
allows to process written text. Refuting the claim that video is good
enough a medium to serve the purpose, we propose to build a representation
that is: editable, queryable, synthesisable and user-friendly---we
define those terms upfront. The issue being functionally and conceptually
linked to that of writing, we study existing writing systems, namely
those in use for vocal languages, those designed and proposed for
SLs, and more spontaneous ways in which SL users put their language
in writing. Observing each paradigm in turn, we move on to propose
a new approach to satisfy our goals of integration in software. We
finally open the prospect of our proposition being used outside of
this restricted scope, as a writing system in itself, and compare
its properties to the other writing systems presented.
\end{abstract}

\section{Motivation and goals}

The main motivation here is to equip Sign Language (SL) with software
and foster implementation as available tools for SL are paradoxically
limited in such digital times. For example, translation assisting
software would help respond to the high demand for accessible content
and information. But equivalent text-to-text software relies on source
and target written forms to work, whereas similar SL support seems
impossible without major revision of the typical user interface.

Video integration seems natural for any kind of SL-related application,
and many SL users make it a preferred storage format, including for
personal notes. Some go as far as to say it serves enough not to need
a written form altogether. The claim is that video provided an acceptable
substitute for textual input, and that technology and its availability
today compensates what might have been a problem up to recently---virtually
everybody now has a camera in their pocket at all times. But when
producing a small mistake in a film take, correcting it requires repeating
it entirely. One might pull up video editing software to cut and stitch
pieces together if lucky enough that the mistake can be resolved in
this way, but this requires more than language skills and leads to
discontinuous signing (cuts, fade-in/out effects, etc.). For better
use and integration in software, we are looking for a way of processing
content comparable to word processors, in other words allowing to
draft and organise discourse by editing pieces of their contents and
to move them around. We call this an \textbf{editable} representation.

Also, when searching through video contents, finding data requires
\emph{both} scanning it and playing parts of it repeatedly, the search
iterations being based on some memory of the contents. It is analogous
to tape recordings when looking for, say, a forgotten rime in a song.
It requires an alternation of winding the tape forward or backward,
and listening from the new position on for at least enough time to
make a decision on the direction of the next search iteration. The
later replacing CDs partially reduced this problem as they used to
(assuming that they are past technology) have direct access to song
(``track'')---and more rarely sub-track ``index''---beginnings.
But nothing outside of the indexed entries could be found without
falling back on some form of manual scan--play process.

Similarly, video indexing with time-tagged labels can help access
key features of a video, but only tagged content can then be found.
There is no possibility of arbitrarily precise data oversight allowing
to target and focus any detail. Indexing contents therefore only partially
solves the general search problem, and moreover requires to be built
beforehand. We are looking for a representation with which data sets
can be directly scanned and searched through, both to remove the need
for separate indexing, and not to restrict searches to what has been
indexed. We call this a \textbf{queryable} representation.

Besides, we have been modelling SL for about a decade, mostly pursuing
the goal of Sign synthesis. Our progress led to AZee, which has produced
increasingly promising results and is now being used to animate avatars
\cite{fil17,filjmd18,nun18}. We come back to it in a later section,
but it is enough for now to say that it is a form of editable and
queryable representation of SL. It also allows rendering anonymous
SL animation, which is a significant advantage over users filming
themselves if contents is to be made public. We call this a \textbf{synthesisable
}representation, and would like to retain this property as much as
possible.

However, AZee is a formal model, only readable if the system is learnt,
and its abstraction layers understood. It is not readable in the sense
that it would be shaped to match human intuition and recognition of
patterns. Given the goal of implementation for human users of the
language, we are aiming at a representation that is also \textbf{user/human-oriented},
facilitating content manipulation through software GUIs.

That said, we also take the three following statements to be true:
\begin{enumerate}
\item storing language content for one's personal use or sharing it between
people is one of the functions of writing (encoding), and implies
reading (decoding);
\item technology and software now have an essential role in manipulating
and disseminating language content, if not indispensable, even for
some social interactions today;
\item \emph{if} there was a writing system used by the community of language
users (shareable, inter-person readable, etc.), \emph{then} they would
rather have their language-related software implement it than be required
to learn a bespoke system for a specific program use.
\end{enumerate}
This naturally brings us close to the question of a writing system
for SL. In this paper, we study writing systems in general, and the
existing or possible parallels for SL.

We do though acknowledge that the purposes of a writing system encompass
a lot more than use in software: note taking, thus ability to handwrite,
institutional/legal archiving... Plus, we have enough hindsight on
writing today to understand that any design choice for a writing system
can yield unforeseen problems a long time afterwards, e.g. dyslexia
is reduced when redrawing letters with more contrastive features.
We therefore wish to keep clear of any claim regarding what SL writing
should be shaping into, or how it should socially be developing, as
much as to feed the scientific discussion on SL writing and digitalisation.

\section{Writing systems}

First, not all human languages have a standard written form known
and put in practice by its users. Second, the languages that do (or
did) are all vocal languages. No SL has such system, known and adopted,
but some forms have been proposed. This section presents the scripts
for vocal languages; we talk about those for signed languages in the
following section.

\subsection{Categories}

It is common to distinguish language writing systems in two groups,
depending on whether its basic symbols encode meaning, or things to
pronounce. In the latter case, a combination of several is normally
required before meaning can emerge. A system of the former group is
called \emph{logographic}; the latter \emph{phonographic}.

Chinese is the most known example of logographic system: ``\includegraphics[height=1.5ex]{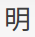}''
is a character meaning ``bright'', ``\includegraphics[height=1.5ex]{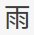}''
another meaning ``rain''. They \emph{happen} to be pronounced\footnote{\label{fn:pinyin}The Chinese phonological transcription system used
here is Hanyu Pinyin.} ``míng'' and ``yǔ'' respectively, but this is not conveyed
by the symbols in a systematic way. The reader must know it to read
them out.

On the other side is Spanish for example, whose Latin alphabet is
used to encode sounds (phonemes) to be pronounced. For example ``p''
encodes the phoneme /p/, and the character string ``pelo'' the phonemic
sequence /pel\textopeno /. It \emph{happens} to mean ``hair'', but
it is not what is captured by the written symbols. The reader needs
to know it to understand the sequence.

\begin{figure}
\begin{centering}
\includegraphics[height=2cm]{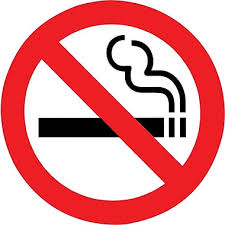}
\par\end{centering}

\caption{\label{fig:no-smoking-pictogram}''No smoking'' pictogram}
\end{figure}

Zooming closer on this rather coarse dichotomy, we see a finer continuum
in the level of linguistic abstraction applied.
\begin{itemize}
\item On the side of symbols encoding meaning, we first reference \textbf{ideographic}
systems. They encode full/fixed meanings with symbols like that in
fig.~\ref{fig:no-smoking-pictogram}, whose interpretation is ``do
not smoke in this area''. But they are not the written form of any
specific oral language, thus we consider them out of our scope.
\item A \textbf{logographic} system composes linguistic elements on a level
that we have seen is at least morphemic, e.g. equivalent to words
like ``rain''.
\end{itemize}

Then, phonographic systems do not necessarily jump straight to encoding
phonemes. There exists various strategies to encode sounds on a sub-morphemic
level.
\begin{itemize}
\item \textbf{Syllabaries} like the Japanese hiragana and katakana have
minimal units that stand for a full syllable weight in time (mora)
each, e.g. the consonant--vowel combinations ``\includegraphics[height=1.5ex]{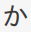}''
for /ka/ in hiragana, ``\includegraphics[height=1.5ex]{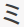}''
for /mi/ in katakana, etc.
\item \textbf{Abugidas} like the Devanagari script used to write Hindi also
encode syllabic rhythm but each syllable character reveals the combined
phonemes, e.g. ``\includegraphics[height=1.5ex]{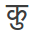}''
/ku/ = base glyph ``\includegraphics[height=1.5ex]{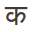}''
/k(a)/ + the diacritic\footnote{A \emph{diacritic} is a mark added to a base glyph in a script.}
for vowel /u/.
\item \textbf{Alphabets} encode sequences of phonemes, with no syllabic
marking. Full alphabets give equal status to vowels and consonants
like in most European languages, whereas \textbf{abjads} mark only
the consonantal base on top of which vowel marking is mainly optional
or partial, usually because imposed by the grammatical role or context,
e.g. Arabic.
\item \textbf{Featural systems} like the Korean Hangul also encode phonemes,
but even reveal the phonetic (articulatory) features to combine to
produce them, e.g. plausive+labial+unvoiced. Noticeably, the Hangul
script also groups the phonemes to reflect syllabic structure.
\end{itemize}
The progression is one of decreasing level of linguistic abstraction,
which more or less follows the order in which respective systems have
tended to appear. The first ones in the history of writing were mostly
logographic, whereas most new ones emerging are phonographic.

\subsection{Implications}

In a purely phonographic system, one:
\begin{itemize}
\item must learn the mapping between sounds (or sound features) and written
symbols;
\item is able to read out (pronounce) any written input and write down anything
heard without understanding;
\item has to know more of the language to make sense of text;
\item uses a relatively small number of symbols all together.
\end{itemize}
Spanish is an almost perfectly phonographic system in this sense.
For example, once you have learnt the following bidirectional mappings:
``m'' is pronounced /m/, ``a'' /a/, ``n'' /n/ and ``o'' /\textopeno /,
you can read ``mano'' to pronounce /man\textopeno / and write the
former when hearing the latter. But the fact that the sequence means
``hand'' is still out of reach until you learn the language.

In a purely logographic system, one would observe symmetric properties:
\begin{itemize}
\item learn direct mappings between written symbols and their meaning;
\item be able to make sense of input text and write down concepts without
speaking the language;
\item knowing the language is required to read text out loud or write down
oral input;
\item a large number of symbols constitute the script.
\end{itemize}
In the Chinese writing system, usually classified as logographic at
first, ``\includegraphics[height=1.5ex]{repl/Chinese-rain}'' is
an example of a non ambiguous logographic character meaning ``rain''.
It is not necessary to know how to pronounce it to interpret it, and
indeed dialects may not all agree on an identical pronunciation for
it. But any capable reader will interpret it correctly.

Note that even in the ``pure'' systems assumed in this section,
nothing precludes ambiguities in text, which are ubiquitous in language.
If phonographic, writing encodes sounds, but similar sounds can bear
multiple and distinct meanings, a phenomenon called \emph{homophony}.
For example in English\footnote{This is a working example of homophony in English for the sake of
reader's understanding, but the properties of this section very poorly
apply to written English, which on the whole is far from a purely
phonographic system. We address this in the next section.}, the sequence ``just'' is composed of 4 ordered graphemes standing
for the respective phonemes /d\textyogh /, /\textturnv /, /s/ and
/t/, together pronounced in sequence as /d\textyogh \textturnv st/
and meaning both ``equitable'' and ``only/merely''. The same written--pronounced
form is interpretable either way, and the ambiguity will remain as
long as context allows.

The logographic counterpart to homophony is \emph{pure synonymy},
i.e. different sounds with undistinguishable meaning. An exclusively
logographic system would write such instances in the same way. However,
such candidates are rather rare as they will likely carry some nuance
at times thus not qualify as identical meanings. Moreover, being pronounced
differently is almost always enough to justify different written forms.
This is what we mean when we oppose ``writing a language'' to ideographic
pictograms, mentioned and discarded further up. A script is the written
form \emph{of a language}, including its various entries and possibilities
for nuances.

\subsection{Neither complete nor fully-exclusive systems\label{section:non-exclusive-systems}}

In general, no system in use possesses all properties of a given class,
in writing and reading directions.

In English, many homophones have different spellings. For instance
``night'' and ``knight'' are both possible written forms of /na\textsci t/,
a phenomenon called \emph{heterograph}y. Conversely, different possible
pronunciations are sometimes written with identical forms, e.g. the
letter sequence ``minute'' in ``this will take a minute of your
time'' (meaning: 60 seconds) vs. ``only a minute fraction of the
total will be lost'' (meaning: very little), respectively pronounced
/m\textsci n\textsci t/ and /ma\textsci nju\textlengthmark t/. They
are called \emph{heteronyms}.

In other writing systems first classified as phonographic:
\begin{itemize}
\item French ``a'' vs. ``à'', both pronounced /a/, mark the difference
between a conjugated auxiliary verb and a preposition;
\item Japanese ``\includegraphics[height=1.5ex]{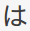}'',
normally standing for /ha/, is actually /wa/ when it is the topic
marker (grammatical function for a particle);
\item German ``du hast'' vs. ``du hasst'', both read /duhast/, are formed
from inflections of the different verbs ``haben'' (to have) and
``hassen'' (to hate) respectively;
\item etc.
\end{itemize}
In all examples above, it is the \emph{meaning} or function that justifies
a distinction in writing or pronunciation, which is not a natural
property of the phonographic approach. Some languages are known to
have a very high grapheme-to-phoneme and phoneme-to-grapheme correspondence
like Finnish or Croatian, but this still often has to exclude things
like loan words. Also, punctuation marks and number digits, part of
the script as much as the letters, encode lots of meaning and very
little or no pronunciation cues. What is more, these scripts consistently
separate words with a space, which we argue is alone a highly functional
(non-phonographic) feature of the script, as nothing allows to know
where the spaces must go on a purely phonemic basis.

The system best classified as logographic, Chinese, also has comparable
irregularities. For example, a character typically has several reading--meaning
pairs. Also, it allows to write pronunciations, which enables transcription
of foreign place names for example. On a lower level, characters are
often themselves composed of pieces including a phonological clue.
For example ``\includegraphics[height=1.5ex]{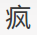}'',
meaning ``crazy, insane'' and pronounced\footnote{See footnote \ref{fn:pinyin}.}
``f\={e}ng'', combines the key ``\includegraphics[height=1.5ex]{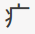}''
(denoting illness) and the character ``\includegraphics[height=1.5ex]{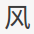}''
(pronounced ``f\={e}ng''). The latter is a pronunciation hint (here
identical to the target) for the whole character; its meaning (``wind'')
is irrelevant. These are not natural features of a logographic system.

Japanese even famously mixes systems right from the start: it involves
the two distinct syllabaries \emph{hiragana} and \emph{katakana},
plus \emph{kanji} characters borrowed from the Chinese script. Used
conventionally, hiragana marks grammatical particles and verb or adjective
endings, katakana loan words and sometimes emphasis, kanji normally
capturing the remaining lexical units. All three systems read out
with the same set of syllables. The first two are phonographic and
encode them directly with a one-to-one mapping; whereas kanji is as
logographic as can be said about written Chinese.

So a mix seems always to be present. The two extreme categories, phonographic
and logographic, are mostly fantasised, all systems instead showing
features of both sides, in variable proportions.

Finally, it must also be noted that a lot is \emph{not} written, and
left to be compensated, even in the most featural phonographic system.
Features like stress can be essential to pronunciation (i.e. mandatory
articulations), but never written. This is true for English. Most
short vowels in Arabic, though one could write them, are left for
the reader to infer from semantic context, recognise from the written
context (surrounding letters and spaces).

\section{SL writing}

As we stated in our introduction, some SL users question the need
or benefit of a writing system for Sign Language. They sometimes argue
that video captures the subtle articulations of the speaker, whereas
any transcription would come with a decrease in precision. Sometimes,
writing is even seen as an intrusive feature of vocal language culture,
if not a threat. We have explained why video should not be considered
sufficient for software processing, but would argue further that it
is simply wrong to equate cameras and pencils. A very good and articulate
multi-fold discussion is provided by Grushkin on this topic \cite{grushkin17},
in the first half of his article.

\subsection{Designed systems}

There have been various attempts to devise SL writing systems, from
personal propositions with local use to others reaching enough popularity
to see some discussion on their potential future as actual writing
systems.

\begin{figure}
\begin{centering}
\begin{tabular}{ccc}
\includegraphics[height=2cm]{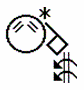} & \includegraphics[height=2cm]{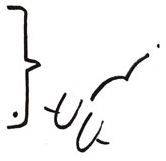} & \includegraphics[height=2cm]{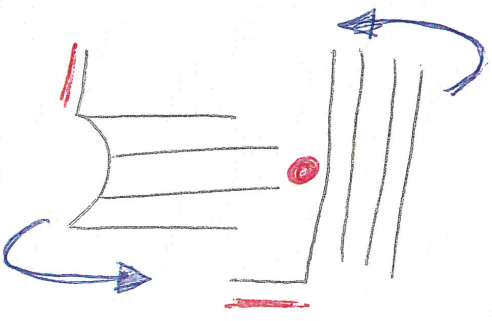}\tabularnewline
(a) SignWriting & (b) ASLwrite & (c) \emph{Signographie}\tabularnewline
\end{tabular}
\par\end{centering}

\caption{\label{fig:planar-systems}SignWriting and similar systems}
\end{figure}

\textbf{SignWriting} \cite{SW-text-book} is by a significant margin
the most visible one. It takes the form of a string of pictures, each
representing a sign and encoding its basic \emph{parameters} as attributed
to Stokoe \cite{sto60}. That is, a hand shape is written for each
active hand, by means of a base symbol filled according to its orientation
in space (facing forward, towards the signer, etc.). Hand location
is given through relative positions of the symbols in the drawing,
movement and contact are shown with arrows and other diacritics. Facial
expression can be specified in a circle representing the head, as
well as shoulder line. Figure~\ref{fig:planar-systems}a gives an
example of a single sign involving one hand with an extended index
configuration, a contact with the temple, a facial expression and
a repeated manual rotation. An interesting feature of this system,
also a voucher of its relative popularity, is that it has been subject
to experiments in deaf classes in several places, in particular to
assess how it could be learnt by early language users. It is also
the only one to have been granted a Unicode block.

\label{systems-like-SW}Other systems have been proposed more or less
following the same underlying principles, e.g. \textbf{Si5s} and its
\textbf{ASLwrite} fork (fig.~\ref{fig:planar-systems}a), which have
been subject to strong promotional efforts. We have also encountered
a system developed by a teacher at INJS\footnote{\emph{Institut National des Jeunes Sourds} = National Institute for
Deaf Youth.}, a Parisian institution for deaf education in Sign Language. She
wished to give students a way to write signs even if they did not
know (or have) a written French equivalent. She called it ``\textbf{signographie}''
(fig.~\ref{fig:planar-systems}b). It is interesting to mention because
it is also taught and used in class by both teachers and students
to support educational communication, though in a different fashion
to the SignWriting experiments. We will be referring to this again
in the next section.

The choice and style of the graphics differ in the systems listed
above, which is relevant when implementing some of the functions of
writing. For example, standard SignWriting requires colouring zones
to show hand orientations, which can be uneasy with a pencil. But
they all otherwise share the same features, capabilities and composition
rules. So in terms of system features and classification, there is
no essential difference between them.

\begin{figure}
\begin{centering}
\includegraphics[width=8cm]{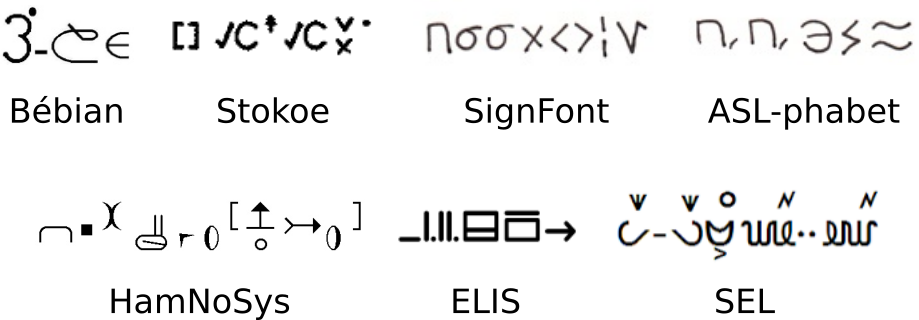}
\par\end{centering}

\caption{\label{fig:linear-systems}Linearised SL writing systems}
\end{figure}

Alternatively, and actually before SignWriting was born, a more linear
approach had been used, which still has representatives. \textbf{Bébian}
\cite{beb1825}, \textbf{Stokoe} \cite{sto60} and \textbf{Friedman}
\cite{fri77} all separated the manual parameters and linearised them
in script, resulting in character sequences, each looking more like
words made of letters and covering what would form a single picture
in a system of the previous group. With the advent of technology,
computers and data processing, more scripts came out falling in this
same category of linearised scripts, given how easier it was to design
fonts, rely on common input devices like keyboards and display TrueType
sequences in word processors. Various such scripts have been proposed
since (see figure~\ref{fig:linear-systems}), intended with more
or less international coverage: the generic \textbf{HamNoSys}\footnote{The Hamburg notation system.}
\cite{pri89,han04}, \textbf{SignFont} \cite{signfont87} and its
follow-up \textbf{ASL-phabet} \cite{aslphabet03} for ASL\footnote{American Sign Language.},
\textbf{ELIS} \cite{meb08} and \textbf{SEL} \cite{les12-SEL} for
LIBRAS\footnote{\emph{Língua brasileira de sinais} = Brazilian Sign Language.}.

HamNoSys\footnote{Hamburg Notation System.} is certainly the most
popular fontified script for SL. It has been used as the main means
of representing signs in academic papers by several scholars, even
doing away with drawings or pictures at times. More impressively
yet, it was implemented, through its XML adaptation SiGML, as the
primary input format to an avatar animation software after the turn
of the millennium \cite{ell04,ell08}. This is a unique property that
was never successfully reached by any other, and which is relevant
to one of our goals here (synthesisable).

\subsubsection*{Observations}

All of the systems mentioned above encode minimal forms to articulate
and combine, though the granularity of the minimal forms are variable.
They show features such as hand shapes (coarse grain), finger bending
(finer grain), hand locations, mouth gestures, and may include more
abstract features like symmetry or repetition instead of duplicated
symbols. In every case it is a description of what must be articulated,
i.e. \emph{form} features. None of them write anything directly mapped
to meaning without an indication of the form. To learn the system
is to learn to articulate the symbols, and it is then possible to
do so without understanding what is read. In this sense they are all
phonographic systems.

\begin{figure}
\begin{centering}
\includegraphics[height=4cm]{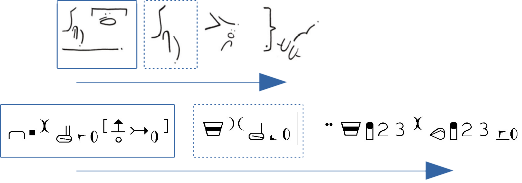}
\par\end{centering}

\caption{\label{fig:unit-sequence}Unit sequences in ASLwrite (top) and HamNoSys
(bottom)}
\end{figure}

Also, they all assume a segmentation on the same level as the one
that linguists use to \emph{gloss} signed input. It is the level usually
called ``lexical'', i.e. of dictionary entries (``signs'') and
other non-lexicalised productions such as classifier placements or
movements. The latter use the signing space in a way that is more
semantically relevant, but they are nonetheless written following
the same composition rules in the scripts. In every system, these
units are stringed one after the other in a linear sequence, as illustrated
in fig.~\ref{fig:unit-sequence}.

At this point we point out that we found no justification for these
design features. They seem mostly taken for granted as a starting
point, whereas we argue they can be questioned, especially in the
light of the wide panel of other known written scripts. For example,
why not a single logographic property? Without implying that things
\emph{should} be different at this point, we will at least be showing
that they \emph{can} in many respects, while thorough exploration
of the alternate paths has not taken place.

Most of the scholarly work we found discussing SL writing systems
either take a phonographic goal for granted without even mentioning
the distinction \cite{mccarty04,kato08}, or do talk about the duality
only to evacuate logography and favour phonography with no compelling
reason to do so \cite{mar00,gui14}. It is probable that this is largely
due to a double cultural influence. Firstly, the systems above originate
from Western cultures with dominant Indo-European languages all written
via phonemic systems juxtaposing whitespace-separated lexical units
in linear text. Secondly, the dominant SL theories in the last five
decades have been inspired by parametric description of signs in Stokoe's
sense, which is rooted in phonology, phoneme inventories and minimal
pairs. Every system presented so far assumes such parametric composition,
and chooses to represent it in some way.

Grushkin \cite{grushkin17} must be one of the rare authors to present
Chinese logography seriously and discuss its benefits and drawbacks.
He even reports on findings telling that deaf Chinese readers have
less difficulty reading logographs than the English do strings of
alphabetical letters, which is a door wide open on logography for
SL writing. Yet somehow he too ends up closing it, advocating what
he calls an ``alphabetical'' (in our terms here, phonographic) paradigm\footnote{As we understand Grushkin's final position in the paper, his support
of an ``alphabetical'' system is in fact one that both favours phonography
over logography and wishes to minimise iconicity in the script. Indeed
he opposes the term ``alphabetical'' to ``iconographic'', which
he uses to mean ``whose phonemic symbols are iconic''. We come back
to this property of iconic symbols further down.}, ultimately to facilitate literacy in English. After such an admirable
plea to equip SL with writing, and so eloquently explaining the need
to empower SL and the Deaf culture with an autonomous system (e.g.
rejecting any sort of glossing, etc.), we find his last call rather
surprising.

At least two major differences with the dominant scripts stand out
though, which we analyse as coming more or less directly from the
difference in the physical channel, because they have to do with simultaneity
and iconicity. The first one comes from the fact that \emph{within}
a lexical sign, phonemic composition is simultaneous and not reducible
to a sequence, like ``just'' is the concatenation of /d\textyogh /+/\textturnv /+/s/+/t/.
This has forced to choose between two strategies, each breaking something
of the alphabetical idea of continuous phonemic sequence. The symbols
to be articulated simultaneously are either:
\begin{itemize}
\item packed in a planar arrangement to form one complex unit, as shown
boxed in the top line of figure~\ref{fig:unit-sequence}---the equivalence
between the sequence of symbols and that of the production is retained,
but the units of the sequence are no more each a minimal phonographic
unit;
\item or linearised, and some form of spacing takes place to separate the
flattened units (bottom line in the figure)---the sequence of symbols
is then to be segmented on two different levels, one of which is no
more an account of the production sequence.
\end{itemize}
Incidentally and likely for the same cultural reasons as above, scripts
generally follow the left-to-right writing direction (arrows in fig.~\ref{fig:unit-sequence}).
An exception is SignWriting, which now prefers a vertical top-down
direction.

The second major difference with common writing systems is about the
symbols themselves. Most of the systems (and all the major ones) have
embedded some iconicity in the graphics, i.e. a resemblance between
the symbols and the way to articulate them. For example, the ``5''
hand shape (flat hand, fingers spread) is drawn \includegraphics[height=2ex]{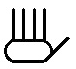}
in HamNoSys, \includegraphics[height=2ex]{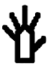} in SignWriting,
\includegraphics[height=2ex]{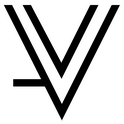} in ASLwrite... They
are iconic of the form to produce, involving 5 countable fingers.

From what we can tell, most writing systems of vocal languages have
actually started this way. Some Chinese logographic characters are
even still reminiscent of that fact, e.g. \includegraphics[height=1.5ex]{repl/Chinese-rain}
(rain). But they have gradually been abstracted, simplified and conventionalised
over time, often giving rise to new or altered meanings, and it is
fair to say that writing systems today are not iconic. Whether it
is natural for a system to lose iconicity over time, and whether or
not there is a different case to be made for Sign Language, we at
least call out this difference for now as the notion always has special
relevance in Sign Language studies.

\subsection{Spontaneous productions}

All the scripts presented in the previous section were designed \emph{systems},
i.e. sets of rules intended to be complete (covering everything deemed
necessary) and consistent (identical events captured with identical
representations). Aside from those developments, in the few years
leading to our present questions, we encountered other uses of pen
and paper aimed at SL representation without relying entirely on dominant
(``foreign'') language support.

First, many SL users taking notes of signed input or preparing signed
speeches resort to graphics to represent the original or intended
signed production. Whether to capture a sign, path, movement, space
location or meaningful relationship between elements of the discourse,
graphical schemes found sufficient to express the production are for
these users naturally preferred as the added cognitive search for
words or phrases in a second language becomes unnecessary.

\begin{figure}
\begin{centering}
\includegraphics[width=0.8\textwidth]{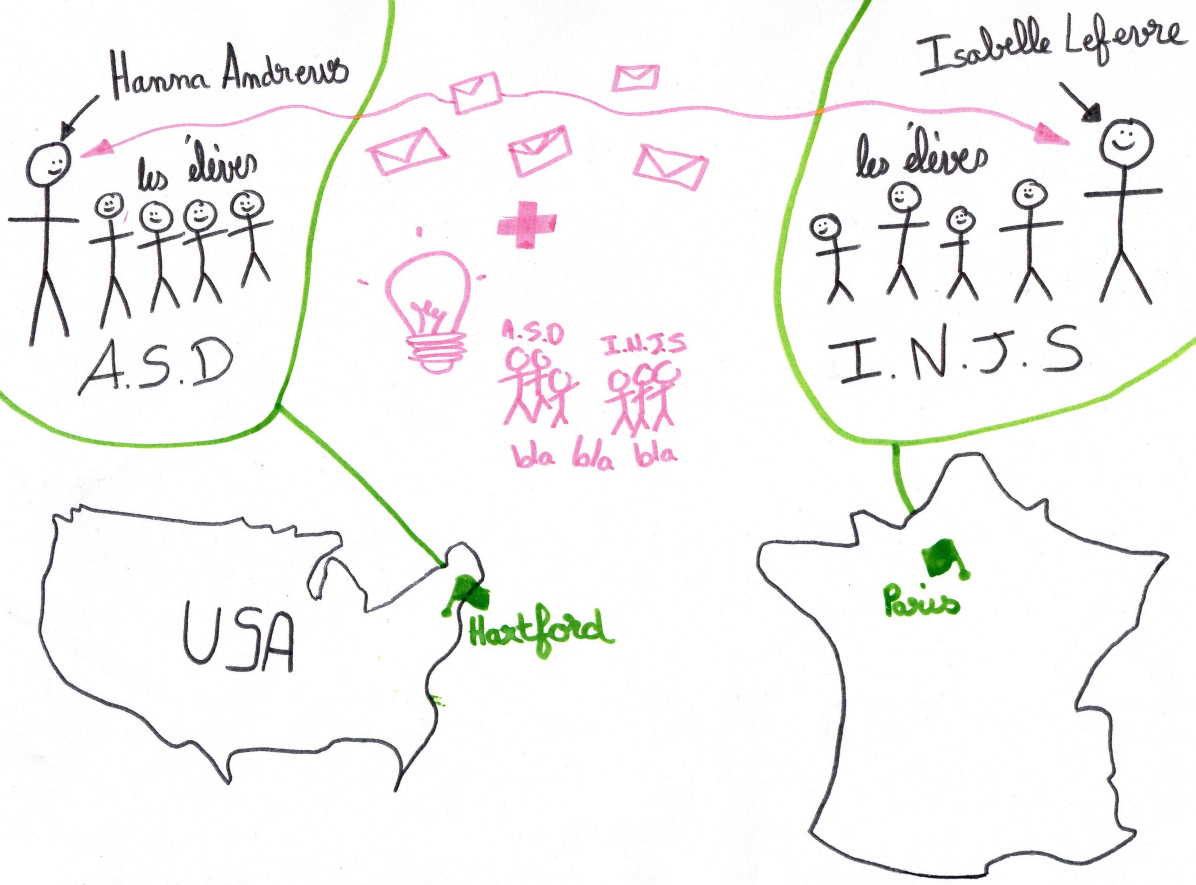}
\par\end{centering}

\caption{\label{fig:VD-INJS-jumelage}Diagram produced by an INJS pupil}
\end{figure}

A second example is in teaching deaf students in signing environments.
Teachers and deaf education experts encourage the use of visual material
for deaf education, even if SL is not the only language used in the
programme. At INJS, we met teachers that have pushed this idea further
than, say, explanatory diagrams to teach new concepts. According to
them, students should be able to turn in work in a written form, the
official written language is a foreign one, and Sign Language is best
captured with drawings. So in addition to \emph{signographie} (see
§\ref{systems-like-SW}), they allow the students to draw SL the way
they feel it should, provided they understand the signing that motivated
the drawing. The school has kindly agreed to share a few of those
productions with us. Figure~\ref{fig:VD-INJS-jumelage} shows one
of the pages of a piece of homework.

\begin{figure}
\begin{centering}
\includegraphics[height=12cm]{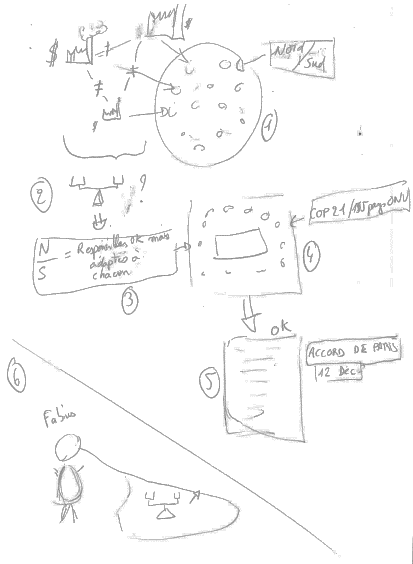}
\par\end{centering}

\caption{\label{fig:VD-COP21}Diagram produced by a French--LSF translator}
\end{figure}

The third use case is that of text-to-Sign translation. Professionals
are taught to draw ``deverbalising diagrams'' as a first step from
input texts to capture all of what must be delivered in SL (the meaning)
while enabling to work without the texts (the source form) afterwards,
so as to produce a semantically equivalent discourse in SL (the target
form) in a way that is detached from the original foreign input. We
have begun to work with LSF\footnote{\emph{Langue des signes française} = French Sign Language.}
professionals on possible software assistance to this deverbalising
task. We give an example of diagram in fig.~\ref{fig:VD-COP21}.

\begin{figure}
\begin{centering}
\includegraphics[width=0.8\textwidth]{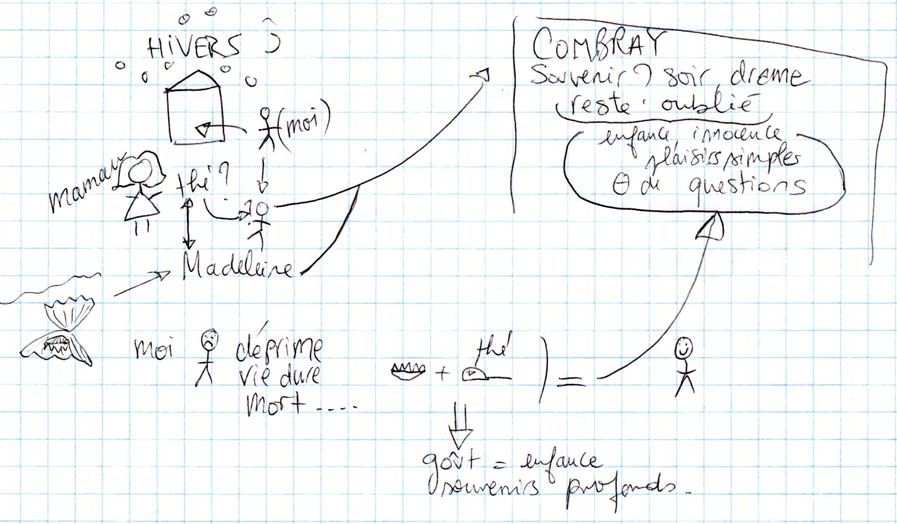}
\par\end{centering}

\caption{\label{fig:VD-semantic-rep}Semantic diagram, hardly SL-specific}
\end{figure}

These diagrams usually lie somewhere on a continuum, already observed
by Athané \cite{ath15}, between:
\begin{itemize}
\item semantic representations, which capture meaning regardless of any
language in which it could be expressed;
\item and what we shall call \emph{verbalising diagrams} (VD) henceforth,
i.e. drawings laid out in such way that they can be followed directly
to produce well-formed SL discourse.
\end{itemize}
Fig.~\ref{fig:VD-semantic-rep} is primarily an example of the first
kind. It looks more like an educational diagram than inspired by SL
particularly. In the case of translation, such a representation will
come from a pure deverbalising effort, and will often be annotated
in a second step with numbers to order pieces in a Sign-logical way
for SL production, though this step is not always easy. The example
figure~\ref{fig:VD-COP21} is closer to the second end of the continuum,
as it produces its own SL-inspired information sequence (the number
annotations only confirm the natural flow of its contents), and every
piece seems to mirror the way to express its meaning in SL. Given
our interest in writing SL specifically, this article will preferably
work with the second kind (VD).

Unfortunately, these are all local or personal productions, usually
intended for short-term use and discarded afterwards. No archive or
data compilation exists of such diagrams whereas, after looking at
the few shared with us, we came to observe much more consistency and
expressiveness than what even their own authors seemed willing to
grant them. We therefore believe science ought to take a better look
at them, and have begun building a corpus of such diagrams to this
effect, aligned with their signed equivalents. The collection involves
various linguistic profiles (e.g. nativeness) and uses (e.g. translation,
note taking, authoring), elicitation material (e.g. text, video) and
genres (e.g. story, definition), etc. It is currently in progress,
and we will present the corpus in detail in subsequent publications.
But the data collected so far already provides discrete examples of
recurrent observations, some of which we wish to expose.

\subsubsection*{Observations}

In this type of graphical representation, meaning plays a major part
in what is written, and on first look a lot more so than form. Let
us first look at the \emph{atomic} level in more detail, i.e. the
smallest, non-breakable symbols populating the diagrams.

What we observe is: all participants drew dogs to mean the pet animal;
none drew what body articulators should do to sign ``dog''. The
same applies for most icons on the collected pages. Without knowing
the language, one can be told---or in this case even guess---what
these symbols mean and understand them regardless of how to sign them,
and nothing from those symbols tells how to sign them for sure.

\begin{figure}
\begin{centering}
\begin{tabular}{ccc}
\includegraphics[height=2cm]{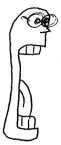} &  & \includegraphics[height=3cm]{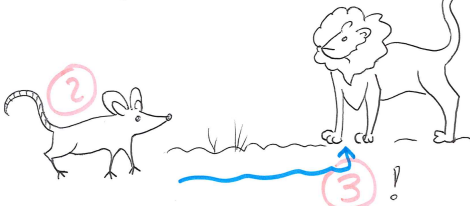}\tabularnewline
(a) & \hspace{1cm} & (b)\tabularnewline
\end{tabular}
\par\end{centering}

\begin{centering}
\begin{tabular}{ccc}
\includegraphics[height=2cm]{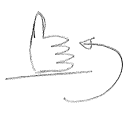} &  & \includegraphics[height=2cm]{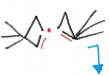}\tabularnewline
(c) & \hspace{1cm} & (d)\tabularnewline
\end{tabular}
\par\end{centering}

\caption{\label{fig:VD-form-over-meaning}VD symbols representing form over
meaning}
\end{figure}

In this sense they make diagrams lean towards the logographic category
of scripts, should they be recognised as such. Although, on this same
atomic level, examples of units representing the signed form (and
not only its meaning) are found in three circumstances:
\begin{enumerate}
\item illustrative/depicting units, e.g. fig.~\ref{fig:VD-form-over-meaning}a,
which represents a jaw drop meaning astonishment and to be reproduced
as a form (a kind of short role shift), or fig.~\ref{fig:VD-form-over-meaning}b,
which represents the path followed by a mouse underground and whose
geometry (wiggling forward then popping out straight up) must be redrawn
in the signing space;
\item \label{list-item:salient-form-abstract-meaning}high salience of form
over ease of representation ratio, e.g. fig.~\ref{fig:VD-form-over-meaning}c
representing the sign for the notion ``the most important, dominant'',
which involves a movement of hand shape ``thumb-up'' (LSF ``1'')
reaching the top of the other ``flat hand''---it is clearly a drawing
of the form to articulate, easier to capture in a drawing than the
rather abstract notion it conveys;
\item the special case where authors knew a phonographic system, and that
it was shared with the potential/intended reader, like in the INJS
environment where the teacher's \emph{signographie} was proposed to
the students for use in their diagrams---for example fig.~\ref{fig:VD-form-over-meaning}d,
encoding the LSF sign for ``result'' in that system, was found used
as a section title.
\end{enumerate}
It is yet to note that all forms in case (1) and many of case (2)
are iconic, hence represent their own meaning in some way, which undermines
the proposition of a phonographic status for these units. Also, out
of the 29 A4-sized pages full of drawings satisfying the premisses
of case (3), we only count 11 instances of \emph{signographie}, which
tells us that even in the case of an available phonographic system,
the preference for meaning is not overturned.

We therefore observe that for atomic elements there is spontaneous
preference of users for logography, though phonography is not avoided
at all costs\footnote{Admittedly, results would be different for borderline cases like writing
poetry, since it is mostly a play on form. The small amount we have
seen of written SL poetry does indeed encode articulated forms in
a more significant proportion (focus on hand configurations, movement
paths and rhythms, etc.)---see page~64 in \cite{lmf15} for an example.}. This is a significant difference with the current offer in designed
systems, which are exclusively phonographic.\label{logography-in-spontaneous-VD}

\begin{figure}
\begin{centering}
\includegraphics[height=2cm]{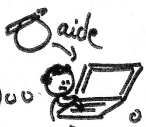}
\par\end{centering}

\caption{\label{fig:VD-relationships}Example of semantic relationship between
two entities}
\end{figure}

Outside of the atomic level, meaning also plays a strong role. Relationships
with various arities\footnote{The arity of a relationship or operator is the number of arguments
it expects. Binary operators have an arity of 2, ternary ones of 3,
etc.} are shown by linking the participating diagram entities with relative
positions, lines and arrows of different styles, sometimes tagged.
They represent semantic relations, often in a way similar to semantic
graphs \cite{sowa08}. Figure~\ref{fig:VD-relationships} is an example
of a directional relation between two people, one helping the other,
represented by an arrow tagged with a (French) word meaning ``help''.
It clearly represents the semantic relationship between the two.

It is tempting to extend the hypothesis of \emph{inclination to meaning}
to the whole process of diagram drawing. But a lot in the end has
to do with form too, again for reasons rooted in iconicity. A high
proportion of the diagrams' layout choices not only serve legibility
purposes or the needs of a 2d projection on the paper. They also perfectly
reflect the spatial arrangements observable in the original signed
discourse if any, and in the later productions when the diagram is
``read''.

\begin{figure}
\begin{centering}
\begin{tabular}{ccc}
\includegraphics[height=3cm]{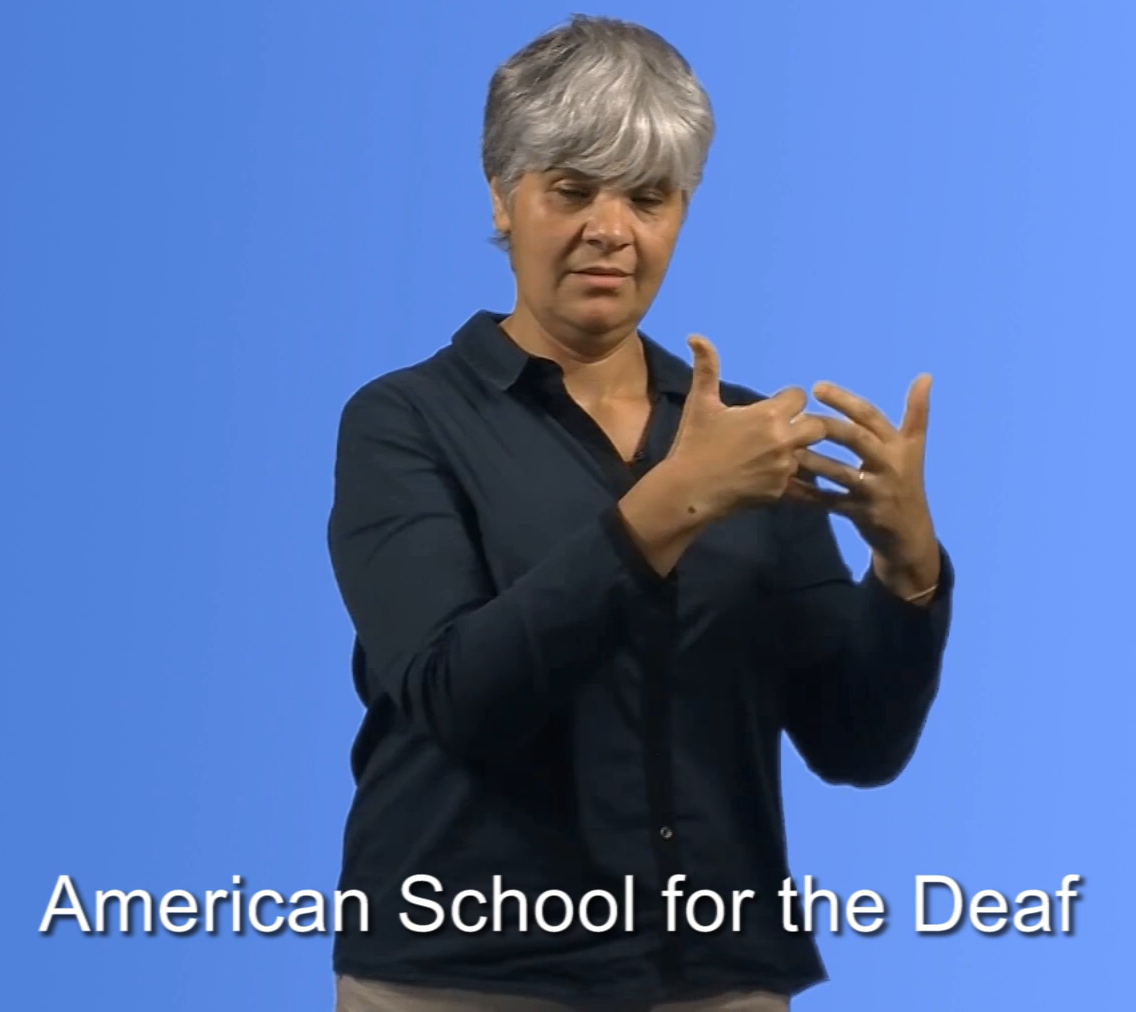} & \includegraphics[height=3cm]{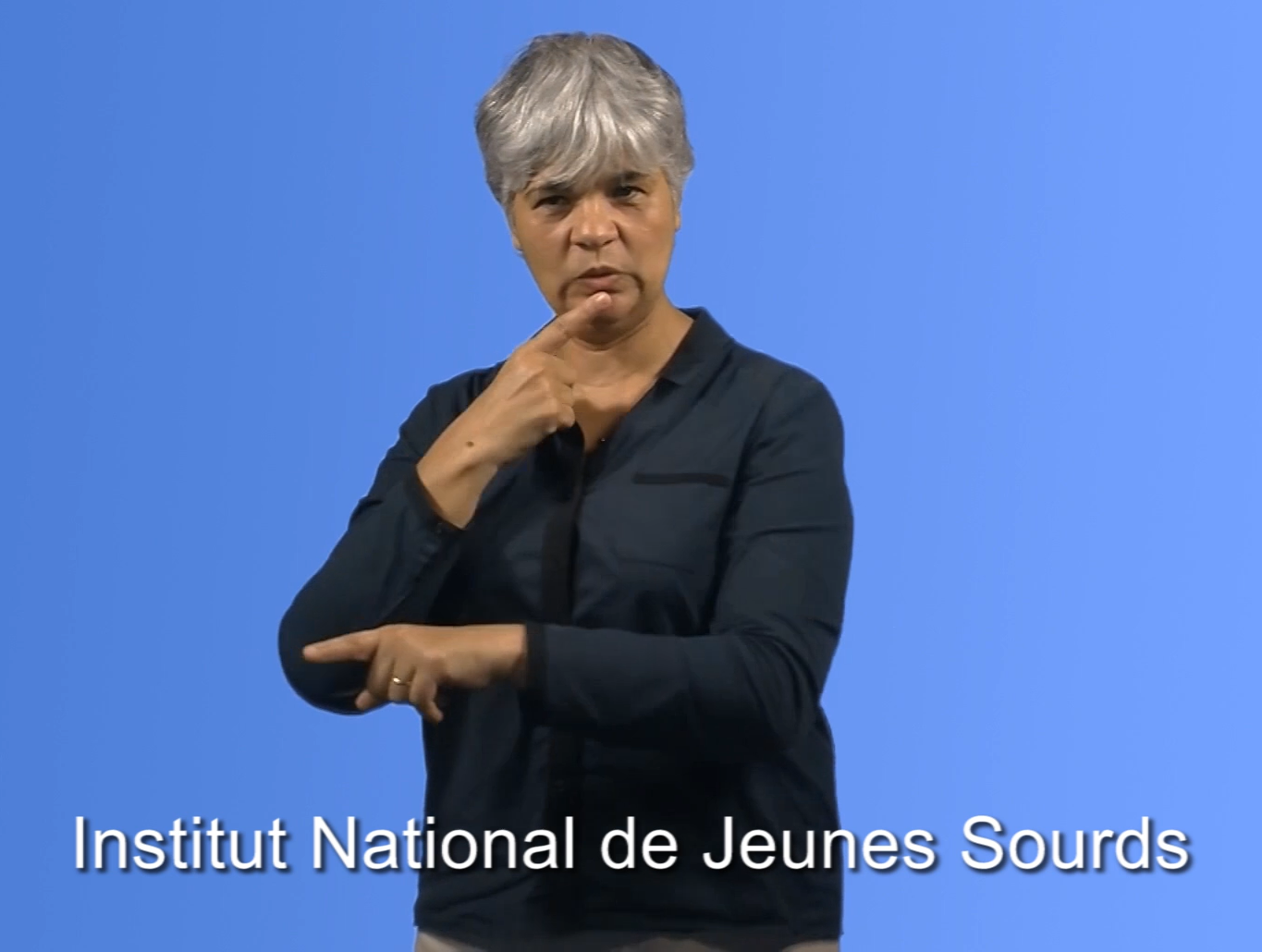} & \includegraphics[height=3cm]{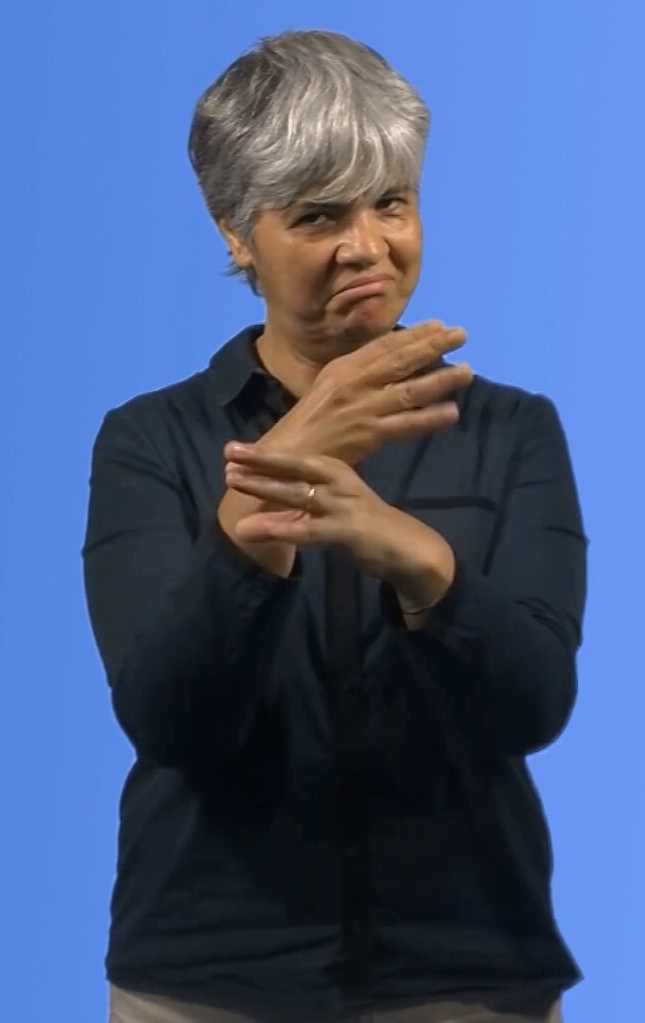}\tabularnewline
(a) & (b) & (c)\tabularnewline
\end{tabular}
\par\end{centering}

\caption{\label{fig:signing-jumelage-Nadia}Snapshots of the SL source of fig.~\ref{fig:VD-INJS-jumelage}}
\end{figure}

\begin{figure}
\begin{centering}
\includegraphics[height=10cm]{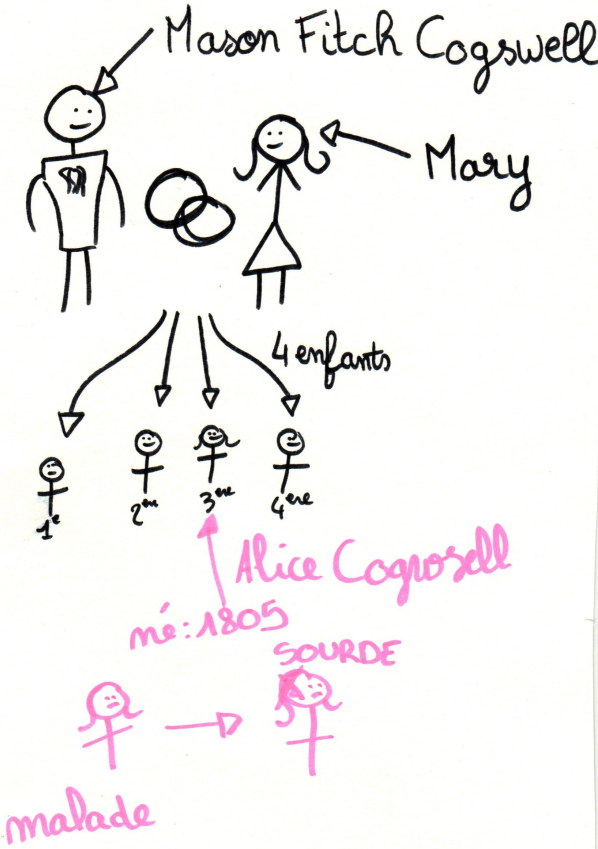}
\par\end{centering}

\caption{\label{fig:VD-family-tree}Diagram layout entirely reproduced in signing
space}
\end{figure}

For example, figure~\vref{fig:VD-INJS-jumelage} is a student's diagram
representing a discourse signed by the INJS teacher who gave the assignment,
about exchange trips between two schools including theirs. Figure~\ref{fig:signing-jumelage-Nadia}
shows three relevant moments of the original video, in order of appearance:
(a) while she anchors the ASD institution\footnote{American School for the Deaf.},
class and teacher on her left; (b) while the French counterpart is
anchored on her right; (c) while she explains that letters were sent
both ways between the two schools. It is clear that the student captured
the exact same layout in the diagram. Similarly, the full page given
in figure~\ref{fig:VD-family-tree} identifies Alice Cogswell (though
misspelt) as the 3rd child of a family of 4, who became deaf after
an illness. A reader with just enough SL will easily see how this
directly maps from the frontal vertical plane of the signing space,
with the scene developing from top (parents) to bottom (the focused
child).

The distinction between meaning and form as the target of the representation
is therefore often difficult to make, and we would argue even nonsensical
in many cases. By definition, iconicity confounds the two. Thus when
it is involved, form is likely identifiable as meaning, and representing
one likely represents the other too.

Meanwhile, whether they capture form, meaning or both, the symbols
used are overwhelmingly iconic, i.e. bear resemblance with what they
represent. In this case there is a clear convergence of almost all
approaches mentioned up to here: they favour iconic symbols in the
script. We consider this an interesting finding: all SL major scripts,
including the designed and the spontaneous ones, use iconic symbols
whereas none can be categorised as such in vocal language scripts.

Contrarily to the scripts presented earlier though, there is no systematic
level of reading where a sequence is to be segmented in ordered units.
Although parts of them happen to be ordered in some places, the diagrams
are essentially two-dimensional. Consequently, it is difficult to
raise the question of a direction of reading, or at least to produce
any conjecture at this point.

\section{Linking to a formal description}

After observing a first set of verbalising diagrams produced by multiple
people, and multiple diagrams for each person, we found recurrent
graphical strategies to capture language components. And something
struck us even more yet: the ease with which those systematic mappings
between graphical forms and meaning could be expressed in AZee. This
section explores these new waters a little deeper.

\subsection{AZee}

AZee is a framework to represent SL in a way that is both linguistically
relevant and formal, in other words unambiguously interpretable by
both humans, e.g. for linguistic accounts of phenomena, and computers,
e.g. for synthesis. We have published about it enough to avoid too
long a diversion here \cite{fil14a,fil16b,fil17}, but this section
summarises the key elements and properties of the model, on which
we build our next proposition.

\begin{figure}
\begin{centering}
\begin{tabular}{ccc}
\includegraphics[height=3cm]{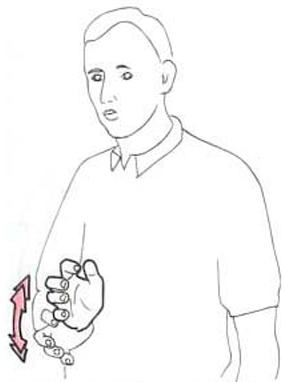} & \includegraphics[height=3cm]{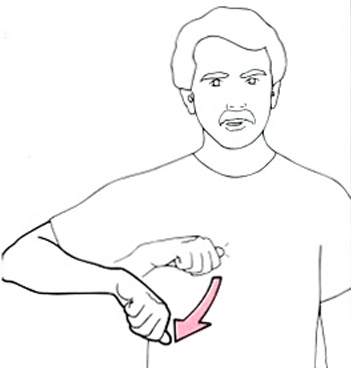} & \includegraphics[height=3cm]{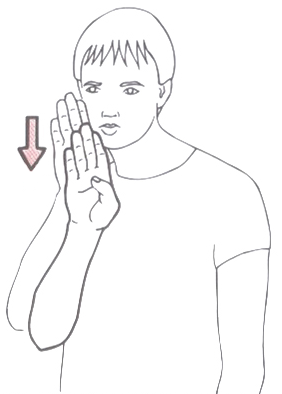}\tabularnewline
(a) ``dog'' & (b) ``scar, surgical cut'' (relocatable) & (c) ``nice, kind, gentle''\tabularnewline
\end{tabular}
\par\end{centering}

\caption{\label{fig:lsf-sign-pictures}LSF sign dictionary pictures \cite{moo86}}
\end{figure}

AZee is an umbrella term for:
\begin{itemize}
\item the general approach to SL description, summarised below, based on
\emph{production rules} and free synchronisation of the whole body
articulator set;
\item a programming language to formalise those rules;
\item the software tool able to compile correctly formed input and generate
\emph{sign scores}, then usable by external software to synthesise
and render animations.
\end{itemize}
The entire approach is built around the duality between observable
form and semantic function, and aimed at bridging them together. To
do so for a given SL, \emph{production rules} are formalised, each
associating necessary forms to an identified semantic function. For
example in LSF, the semantic function ``dog'' is associated the
form shown in fig.~\ref{fig:lsf-sign-pictures}a. This association
allows to define a production rule which when applied generates a
\emph{signing score} specifying the gestures/movements (forms) to
articulate to mean ``dog''. Notably, all of this is done with no
level distinction such as morphology vs. lexicon. The fact that the
result of ``dog'' is often categorised as a lexical production is
irrelevant as far as the model is concerned.

A rule can be parametrised if parts of its form depend on an interpretable
piece of context. The meaning ``surgical cut, scar between points
$P_{1}$ and $P_{2}$ on the body'' is associated a form which includes
a movement between $P_{1}$ and $P_{2}$. Fig.~\ref{fig:lsf-sign-pictures}b
gives an example with both points on the abdomen. The rule is parametrised
with the two point arguments accordingly, and specifies a resulting
score whose description depends on those arguments. Later applying
the rule to two (meaningful and only given in context) points of the
body will automatically generate the appropriate form, in accordance
with those points.

\begin{figure}
\begin{centering}
\includegraphics[height=1.5cm]{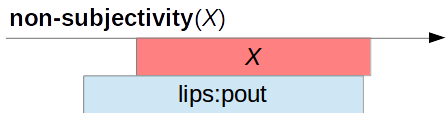}
\par\end{centering}

\caption{\label{fig:boxes-non-subj}Signing score produced by the AZee expression
``non-subjectivity($X$)''}
\end{figure}

\begin{figure}
\begin{centering}
\includegraphics[height=1.5cm]{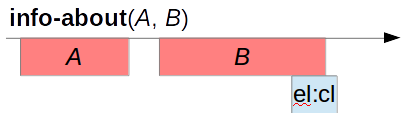}
\par\end{centering}

\begin{centering}
{\small{}$A$ and $B$ must be provided as parameters}
\par\end{centering}{\small \par}

\begin{centering}
{\small{}``el:cl'' encodes an eye blink}
\par\end{centering}{\small \par}

\caption{\label{fig:boxes-info-about}Signing score produced by the AZee expression
``info-about($A$, $B$)''}
\end{figure}

\begin{figure}
\begin{centering}
\includegraphics[height=2cm]{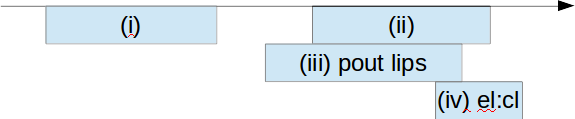}
\par\end{centering}

\begin{centering}
{\small{}(i) contains the form in fig.~\ref{fig:lsf-sign-pictures}a}
\par\end{centering}{\small \par}

\begin{centering}
{\small{}(ii) contains the form in fig.~\ref{fig:lsf-sign-pictures}c}
\par\end{centering}{\small \par}

\begin{centering}
{\small{}(iii) lip pout generated by ``non-subjectivity''}
\par\end{centering}{\small \par}

\begin{centering}
{\small{}(iv) eye blink generated by ``info-about''}
\par\end{centering}{\small \par}

\caption{\label{fig:signing-score-E1}Signing score resulting from (E1)}
\end{figure}

Parameters can be of any type defined by AZee (geometric vector, numerical,
left/right body side...). In particular, rules can be parametrised
with signing score arguments to allow recursive use of production
rule applications as arguments for others. For example, the semantic
function ``it is generally agreed that $X$'' produces a signing
score whose specification is a mouth gesture (lip pout) synchronised
with $X$ with a time offset (see fig.~\ref{fig:boxes-non-subj}).
We have elsewhere called this function ``non-subjectivity'' \cite{fil17}.
This all together allows to generate functional expressions, which
when evaluated produce complex utterances nesting scores in one another.
For example, expression (E1) below combines 4 rules and evaluates
to the signing score given in fig.~\ref{fig:signing-score-E1}.
\begin{description}
\item [{E1}] info-about(dog(), non-subjectivity(nice-kind()))
\end{description}
In the expression:
\begin{itemize}
\item ``info-about($A$, $B$)'' means ``$B$ is the (focused) information
about $A$'' and produces the score in fig.~\ref{fig:boxes-info-about};
\item ``nice-kind'' has a self-explanatory name and produces the form
shown in fig.~\ref{fig:lsf-sign-pictures}c when applied.
\end{itemize}

For legibility, AZee expressions are often represented as trees where
child nodes are the nested expressions used as arguments of the parent
in the right order. For (E1):

\Tree [.info-about dog [.non-subjectivity nice-kind ] ]

It was observed that recursively combined rules produce scores that
can be interpreted as a whole as the combination of their respective
interpretations. For example, the reader might already have interpreted
(E1) as meaning ``dogs are {[}generally thought to be{]} nice'',
which is what one interprets from the production scored in fig.~\ref{fig:signing-score-E1}
(to the extent that such constructed examples allow out of context).
Therefore while AZee trees look like syntactic trees, they are rather
comparable to semantic representations because unlike syntactic trees,
every rule node carries meaning (or would not exist at all).

\subsection{Bridging VD to AZee}

The quest for AZee production rules in LSF has now been going on for
a few years. And with no claim of it being complete yet, our current
set usually allows to almost cover monologues of the informative type
such as news reports. This makes us consider the approach as worth
pursuing and the state of the rule set, if not definite, solid enough
to entrust.

Now as we hinted while introducing the section, patterns in the collected
diagrams were found, which could easily be expressed with identified
AZee rules. Let us list a few. We do not give counts or statistics
as they will not be meaningful at this stage, but we do give a few
of the clear qualitative tendencies. We have already mentioned trivial
examples while discussing the tendency for logography on what we called
the atomic level. For example, the drawing of a dog corresponds to
the rule ``dog''. But other patterns arise on higher levels.

An example of a repeated pattern is the use of \emph{context bars}:
a piece of the drawing $C$ (which can itself combine multiple pieces)
is ``followed'' by\footnote{\label{fn:assumed-direction-of reading-in-VD}This locally assumes
a direction of reading in the diagrams. In all cases observed, this
was top-down and left-to-right, which corresponds to the culture of
the informants' environment (France).} a straight line separating it from a second piece $F$. The already
shown fig.~\ref{fig:VD-COP21} contains an example of this feature.
The overall interpretation is that information $F$ is focused, but
given in the context of $C$. In the SL equivalent (source or result),
the portion corresponding to $C$ is always signed first, and followed
by $F$.

We often make the same interpretation of the following \emph{colour
change} pattern: a scene $C$ is drawn, on top of which a piece of
drawing $F$ is superimposed in a different colour, with $F$ being
signed after $C$ in the SL production. Fig.~\ref{fig:VD-form-over-meaning}b
is an example of such colour change. The two animal entities set up
a context scene (initial positions) in which the blue arrow (movement
of the mouse) is the focused event. Fig.~\ref{fig:VD-INJS-jumelage}
exhibits the same pattern on a larger scale: the exchange of letters
drawn in red is the focused information, and occurs between two sides
set up as context by the rest of the drawing. And the same goes for
fig.~\ref{fig:VD-family-tree}. Of course this is limited to cases
where colour was available; some informants have indeed chosen to
use a single colour. Also, we found other uses of colour change, but
focusing/highlighting a piece $F$ in a contextualising scene $C$
is a frequent one. The noteworthy thing here is that both colour highlighting
and context bars have a direct mapping to the one-rule AZee expression
``context($C$,~$F$)'', which means ``$F$ in context $C$''.

\begin{figure}
\begin{centering}
\includegraphics[height=2cm]{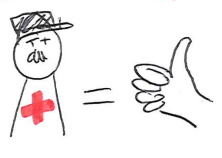}
\par\end{centering}

\caption{\label{fig:VD-equal-sign}Use of the equal sign for function ``info-about''
(in this particular case: ``Fidel Castro is well'')}
\end{figure}

\begin{figure}
\begin{centering}
\includegraphics[height=3cm]{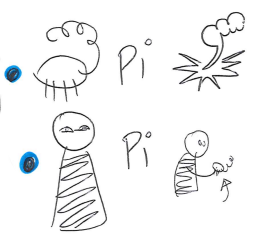}
\par\end{centering}

\caption{\label{fig:VD-bullet-list}Use of a bullet list with function ``each-of''}
\end{figure}

Another repeated pattern is when two pieces $A$ and $B$ are drawn
side by side with more or less similar sizes and an equal sign (``='')
in between, as shown in fig.~\ref{fig:VD-equal-sign}. These instances
match the AZee expression ``info-about($A$,~$B$)'', already introduced
above. A graphical pattern using \emph{bullet lists} (e.g. fig.~\ref{fig:VD-bullet-list})
also emerged for exhaustive (closed) enumerations of simultaneously
true/applicable items, which has its AZee rule ``each-of''...

Such regularities keep surfacing in the diagrams. We must investigate
them further and try our observations statistically on the full corpus
to come. It will exceed 200 drawings, and keep growing as more informants
might still wish to contribute afterwards. But to summarise, at this
point we make the two following observations:
\begin{itemize}
\item spontaneous verbalising diagrams and AZee trees are both in essence
close to semantic representations shaped for production in the target
SL;
\item they share common structuring and composition elements to represent
the meaning.
\end{itemize}
This apparent proximity made us want to explore the possibility of
a bridge between the two types of representation. Let us first acknowledge
an interesting symmetry between them.

First, even in the form of trees, AZee expressions are of a mathematical/formal
nature, in other words friendly only to those familiar with the model,
not human-oriented drawings easy to draw and decipher. On the contrary,
VDs are graphical objects spontaneously used by many who wish to put
SL in some form of writing. Thus they can only be viewed as accessible
to humans, and considering our goal, as a way to ease the interface
between users and software.

Conversely, VDs do not provide full access to the forms to produce
to read them. \label{AZee-is-synthesisable}Whereas, unlike formal
semantic representations like conceptual dependency \cite{schank72},
semantic graphs \cite{sowa08} or more theoretical concepts like ``interlingua''
which are intended to be detached from any specific language, every
AZee expression produces definite forms. That is to say that given
any representation in AZee, a computer program can automatically generate
the corresponding sign score. In the terms defined in our introduction,
it is synthesisable, which is a desired property for our editable
form.

This property comes from the fact that when building an AZee rule
set, any abstraction of observed signed forms behind semantically
relevant rules is done by embedding a link to the factorised forms
inside the abstracted rule. So the forms are hidden in subsequent
expressions invoking the rule, but retrievable to produce a result.
We have elsewhere called this building ``from the target and back''
\cite{fil-fal-arXiv}.

Looking to translate forward from VD to AZee would currently be an
ill-defined task as we have seen that VD is all but non-standard.
Instead, we propose to follow the idea above and build a graphical
tool kit \emph{back from} AZee, in other words to define graphics,
icons, symbol layouts, etc. for AZee rules and structures we already
know exist for sure. Then like no AZee rule exists without an associated
interpretation and form description, no graphics will be made available
without an associated AZee equivalent---which itself comes with meaning
and forms to produce. So the Ariadne's thread leading to the ultimate
forms will always be preserved.

\subsection{A new type of editable SL representation}

\begin{figure}
\begin{centering}
\begin{tabular}{cccc}
\includegraphics[height=3ex]{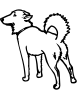} & \includegraphics[height=3ex]{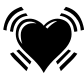} & \includegraphics[height=14mm]{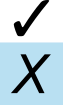} & \includegraphics[height=8mm]{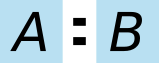}\tabularnewline
(a) & (b) & (c) & (d)\tabularnewline
\end{tabular}
\par\end{centering}

\caption{\label{fig:AZVD-from-AZee}Building a graphical script back from AZee}
\end{figure}

\begin{figure}
\begin{centering}
\includegraphics[height=14mm]{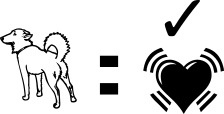}
\par\end{centering}

\caption{\label{fig:AZVD-E1}Proposed script for (E1)}
\end{figure}

The simplest plan to start building back from AZee without losing
coverage is to assign a graphical form to every possible node of an
AZee tree. Such node is of either kind below:
\begin{itemize}
\item a \emph{rule node} referring to a named production rule: this is the
most expected case, and indeed that of all nodes in example (E1);
\item a \emph{native node} containing an AZee expression to build or reference
basic/geometric objects like a numerical value or a point in space
or on the body: for example, two native leaf nodes would come under
``scar, surgical cut'' as arguments for the parent rule node.
\end{itemize}
On top of this, here are two additional recommendations we wish to
follow:
\begin{itemize}
\item the first tentative graphics for an AZee target should be close to
any spontaneous regularity already observed in the diagrams, as it
should maximise intuitive reading and minimise the difference with
VD;
\item for atomic symbols or icons, prefer Unicode characters, as their code
points are secured across future fonts and systems, and many are available,
all the more so as emojis are entering the standard set.
\end{itemize}
For example, the rule ``dog'' can be assigned the atomic symbol
(a) in fig.~\ref{fig:AZVD-from-AZee}, and ``nice-kind'' the symbol
(b). They are both standard characters, with respective code points
U+1F415 and U+1F493. Furthermore, rule ``non-subjectivity'' can
be graphically represented with the addition of, say, a tick mark
over its argument $X$ as shown in (c), and template ``info-about($A$,
$B$)'' with the arrangement (d) observed in our data. Accounting
for the recursive nature of AZee expressions, diagrams for $X$, $A$
and $B$ can themselves be diagrams of complex expressions, which
creates the possibility of recursive diagrams. Combining these graphical
operations, expression (E1) would then be encoded as shown in figure~\ref{fig:AZVD-E1}.

This is very similar to the way arithmetic expressions are written
in the standard math script, as their elements are operators and atomic
elements nested in one another to form one recursive structure. That
is, figure~\ref{fig:AZVD-E1} is to (E1) as the written expression
$\sqrt{\frac{a+b}{c}}$ is to the recursive structure below, representing
operator and argument nesting levels:

\Tree [.square-root [.fraction [.sum $a$ $b$ ] $c$ ] ]

More graphics can later be defined for larger AZee templates, i.e.
AZee sub-trees of more than a single node, for constructions that
are frequent and semantically salient enough. For example, figures~\ref{fig:signing-jumelage-Nadia}a
and \ref{fig:signing-jumelage-Nadia}b anchor a discourse $A$ on
the left-hand side ($A$ is signed with hands and shoulder line turned
to the left), then another $B$ on the right. This is a frequent form
for comparison or opposition of some sort between $A$ and $B$, for
which a possible AZee tree is\footnote{``Lssp'' and ``Rssp'' are native AZee nodes referencing two points
in signing space in front of the signer, to the left and right of
the centre respectively.}:

\Tree [.each-of [.localised-discourse Lssp $A$ ] [.localised-discourse Rssp $B$ ] ]

\begin{figure}
\begin{centering}
\includegraphics[height=10mm]{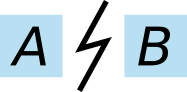}
\par\end{centering}

\caption{\label{fig:VD-from-AZee-comp-opp}Left--right comparison/opposition
between two $A$ and $B$}
\end{figure}

This template could be abstracted as a whole into a binary graphic
operator to further abstract the combination into something that directly
makes sense to the users like ``opposition of $A$ and $B$''. For
example, we could represent it as shown in figure~\ref{fig:VD-from-AZee-comp-opp}.

Following the same principle, we can provide more and more useful
abstractions of AZee templates into compact graphical representations.
At any rate, we see how this develops into a planar (2d) representation
with a recursive structure, equivalent in content to an AZee expression
but more helpful in appearance to human apprehension.

This completes the list of properties initially expressed as our goals
for use in software. Namely, our proposition is:
\begin{itemize}
\item editable, as pieces can be copy-pasted and edited like formulae in
many math editors;
\item queryable, as the input structure can be parsed by a computer and
its contents searched;
\item human-readable, because it is graphical, and the graphics chosen to
be iconic and resemble what humans already produce spontaneously;
\item synthesisable, as they are equivalent to synthesisable AZee expressions,
as explained in §\ref{AZee-is-synthesisable}.
\end{itemize}
The prospect opened here is that of SL editing software that enables
saving, modifying and sharing content, enables quick search functionalities
like the now ubiquitous ``Ctrl+F'' shortcut in applications, and
is linkable to signing avatars for regular (oral) SL rendering.

\section{Prospect for writing SL}

As we said right at the beginning, we expect users of a Sign Language
to want software input to match their own written practice if they
have one some day. Thus we now want to consider our above proposition
outside of its intended scope of software integration, and characterise
its properties in the midst of the scripts mentioned so far.

\begin{figure}
\begin{centering}
\includegraphics[width=1\textwidth]{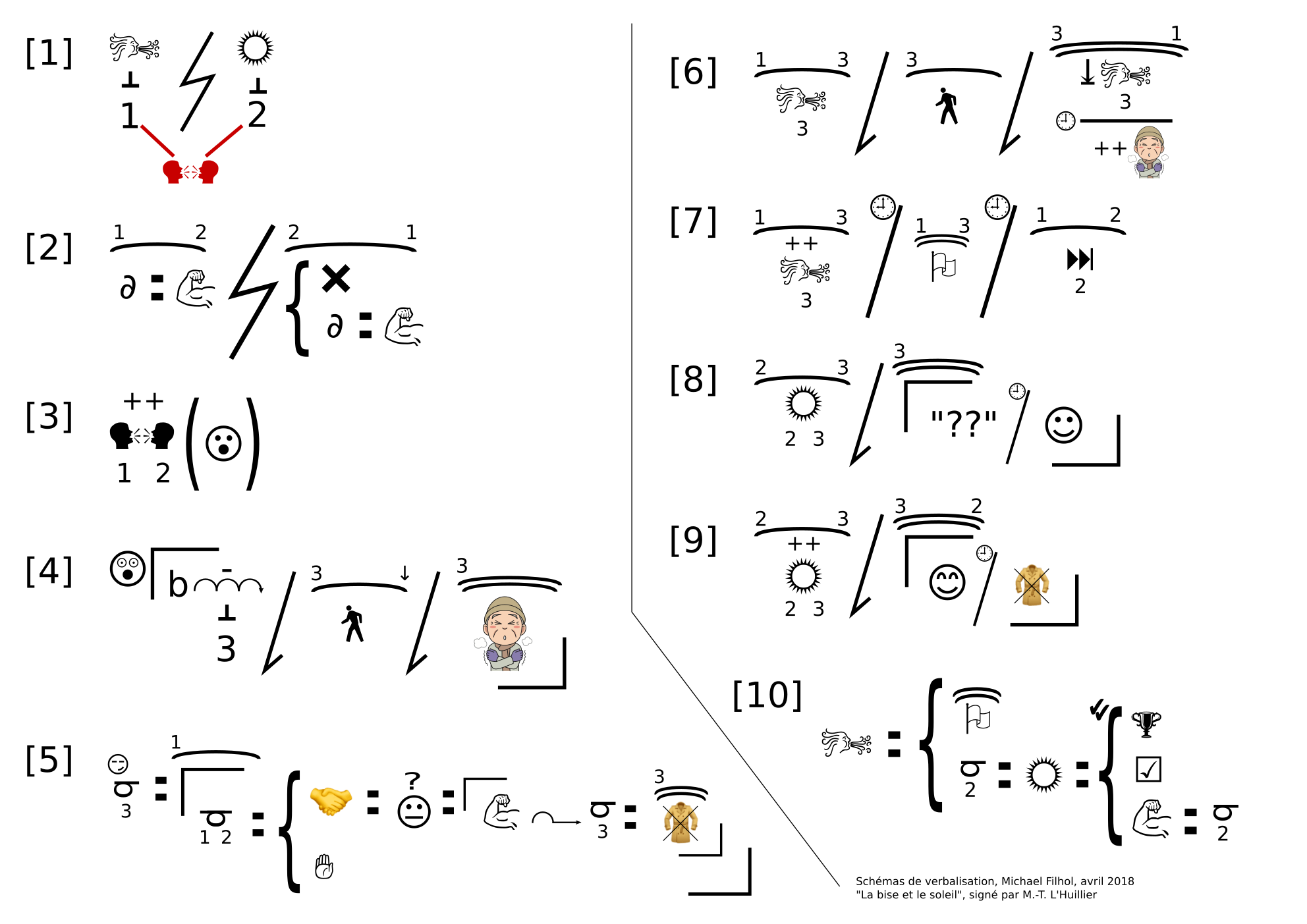}
\par\end{centering}

\caption{\label{fig:VD-from-AZee-BiseEtSoleil}Pieces of a full story}
\end{figure}

For a better grasp of what such proposition would turn out like, we
have written the full AZee tree for a signed version of the short
story \emph{La bise et le soleil}\footnote{In English: ``The Bise {[}North/cold wind{]} and the Sun''. The
version used here is viewable at \texttt{https://atlas.limsi.fr/?tab=LNT}
(click on the ``LSF'' entry).}, and encoded each node with a symbol as proposed, with only a few
approximations or assumptions when AZee coverage was still limited
(unexplored language phenomena). Figure~\ref{fig:VD-from-AZee-BiseEtSoleil}
shows the result for the ten pieces of the overall 80-second performance.

The actual look of the writing shown here here is outside of our present
consideration. The symbols and line styles put out are all only tentative,
if not mere dummies to instantiate our theoretical approach as it
can develop. We even leave out explaining the encoded tree, since
our interest at this point is only to characterise the type of script
we are dealing with, and some of its properties.

The first question of logography vs. phonography proves tricky if
we state the fact that except for native nodes, every glyph refers
to a form--meaning association. In other words then, what is written
is not \emph{either} a form or a meaning, but necessarily both. For
example, the tick mark suggested in fig.~\ref{fig:AZVD-from-AZee}c
maps to both the ``generally found true'' interpretation and the
lip pout form, jointly. From this angle, we would have to question
the logo--phono dichotomy at its roots. But while symbols could in
theory be arbitrary, they never appear to be so in spontaneous VD,
rather as we noticed, they systematically appear as iconic of something.
So if we keep supporting inclusion of observed VD symbols in our script,
we are led to prefer \textbf{iconic} symbols over abstract ones. As
we have already mentioned, this compares to many proposed systems
for SL, and similarly contrasts with the writing systems used in the
vocal world.

Then, the question is raised of what they should be iconic of, and
the two-way characterisation of the script makes sense again if we
consider the iconic references chosen for its symbols. The nature
and structure of the script seems to invalidate the initial question,
but the iconicity put into its glyphs reenables it. For example, the
tick mark proposed in figure~\ref{fig:AZVD-from-AZee}c is more iconic
of its meaning, but could instead be written as lips to indicate the
associated form. We have already touched on this while trying to characterise
VD as a type of script in section~\ref{logography-in-spontaneous-VD}.
We have seen that in most of spontaneous VD, glyphs were iconic of
the meaning, while they would still occasionally refer to form. For
instance it is possible that lips be favoured over the tick mark,
as the choice can fall under list item~\vref{list-item:salient-form-abstract-meaning}.
Whatever the ultimate choices, it seems unlikely that such system
grow into the 100\% phonographic state of the other designed propositions.
By analogy with VD, we rather predict a mix with \textbf{logographic}
references, if not even a dominant number of them. As we have shown
in §\ref{section:non-exclusive-systems}, unbalanced mixes of phonography
with logography are a typical feature of the world's writing systems.
This time our system would therefore be no different.

The issue of directionality must also be revisited, because no linear
layout is assumed to begin with. We do note some directional reading
of operators\footnote{Also see footnote~\ref{fn:assumed-direction-of reading-in-VD}.}
with arity greater than one and whose reversal would seem unnatural
(depending on the subject's culture and literacy) or ambiguous (an
arrow can be reversed, but not an equal sign). But they are local
issues related to specific operators used in the script, not a property
of the script design itself which would organise, say, a whole page
systematically. Instead, it is 2-dimensional, or \textbf{planar}.
What is more, it gives an account of the embeddings of the written
discourse pieces nested in one another, which allows to qualify the
system as inherently \textbf{recursive}. This scripting layout is
identical to that of the preferred mathematical script, which lays
out a planar recursive structure, including local directionality for
specific operations\footnote{Interestingly, math script is also dependant on the surrounding culture
and literacy. Arabic vs. European math scripts are often reversed
where a direction of reading is required.}. So these last two properties would be rather new if ported to a
natural language script, but are not alien to scripting, handwriting
or reading practice.

\section{Conclusion}

In this paper, we created an editable, queryable, human-readable and
synthesisable representation to implement in SL-related software.
It builds on two complementary grounds:
\begin{itemize}
\item the spontaneous handwriting of SL users in the form of verbalising
diagrams, which we are building a corpus of and finding patterns in;
\item the formal AZee framework, which we have been promoting for a few
years, especially in the SL processing community.
\end{itemize}
The point of sourcing from VD is to tap into already existing practice,
increase intuitive use and human apprehension of the system, and to
inspire the graphical layout of the resulting diagrams. The point
of AZee is to take advantage of the formal background, and cancel
the subjectivity that is intrinsic to VD's personal and non-standardised
practice. By building backwards from the formal base, we also guarantee
synthesisable input. We now wish to try out our proposition by implementing
and demonstrating a simple software editor.

In a final section, we have opened the prospect of seeing our proposition
used outside of the restricted scope of software interaction, in particular
as a writing system. We have compared its properties to the other
writing paradigms presented, existing for vocal languages or designed
for SLs. We have shown that it fits key characteristics of writing
systems (mixing phonography and logography), while exhibiting more
or less new properties (iconic, planar, recursive). We hope that after
a few developments, SL users will eventually be tempted to test it
as it will at least help us shape up what is implemented in software,
but also feed the important and difficult reflection on SL writing
and literacy. Whether or not it actually proves robust or adaptable
to all uses signers will require of a writing system in the future
is of course not yet possible to tell.

\section*{Acknowledgements}

We express warm thanks to INJS (Paris, France) and Interpretis (Toulouse,
France) for sharing their pupils' and translator's diagram productions,
as well as for the time spent discussing their use and purpose in
their profession.

\bibliographystyle{plain}
\bibliography{bibliography}

\end{document}